\documentclass[letterpaper]{article} 
\usepackage[preprint]{twocolstyle}  
\usepackage[hyphens]{url}  
\usepackage{graphicx} 
\usepackage{natbib}  
\usepackage{caption} 
\usepackage{algorithm}
\usepackage{algorithmic}

\usepackage{newfloat}
\usepackage{listings}
\DeclareCaptionStyle{ruled}{labelfont=normalfont,labelsep=colon,strut=off} 
\floatstyle{ruled}
\newfloat{listing}{tb}{lst}{}
\floatname{listing}{Listing}

\usepackage{booktabs}
\usepackage{amsmath}   
\usepackage{amssymb}   

\title{ToolVision: Learning When and How to Use Visual Tools\\ with Capability-Aligned Supervision}
\author{
    Delin Mao\equalcontrib\textsuperscript{\rm 1},
    Chenghao Sun\equalcontrib\textsuperscript{\rm 2},
    Jingwei Song\textsuperscript{\rm 1},
    Chishui Chen\textsuperscript{\rm 3},
    Linfeng Zhang\textsuperscript{\rm 1}\corresponding
}
\affiliations{
    \textsuperscript{\rm 1}Shanghai Jiao Tong University\\
    \textsuperscript{\rm 2}University of Chinese Academy of Sciences\\
    \textsuperscript{\rm 3}Fudan University
}

\begin{document}

\maketitle

\begin{abstract}
Thinking with images allows a multimodal model to compensate for limited
perception by invoking visual tools through code. Yet the prevailing
SFT-then-RL recipe creates a different supervision misalignment at each stage. SFT
is expected to teach how to use tools, but trajectories from stronger teachers
may succeed through perceptual capabilities that a smaller student cannot
reliably reproduce or exploit, causing the student to imitate tool-call
patterns without learning how to make them useful. RL is expected to teach
when to use tools, but outcome-only rewards make fallible tool execution a
liability and suppress tool use, whereas a blanket bonus for every correct
tool-using trajectory encourages valid but ineffective operations. To address
these two misalignments, we introduce ToolVision. During SFT, a multi-agent
pipeline explores candidate trajectories, and a committee including
student-scale models scores stepwise evidence gain to rank and prune the
search branches. Only
successfully executed trajectories with correct final answers are retained for
SFT. Before RL, ToolVision 
 compares the learner's performance with and without tools, then rewards
successful tool use only on questions where tools provide a clear benefit.
Both signals are constructed automatically from public task data without
additional human annotations of tool use or necessity. ToolVision-8B improves
over its base on all seven main benchmarks, surpasses Thyme-7B,
CodeVision-8B, and CodeDance-7B on all three high-resolution benchmarks, and
outperforms Qwen3-VL-32B-Thinking on V\textsuperscript{*} and HRBench~8K.
We will publicly release the datasets and source code.
\end{abstract}

\begin{figure}[t]
\centering
\includegraphics[width=0.98\columnwidth]{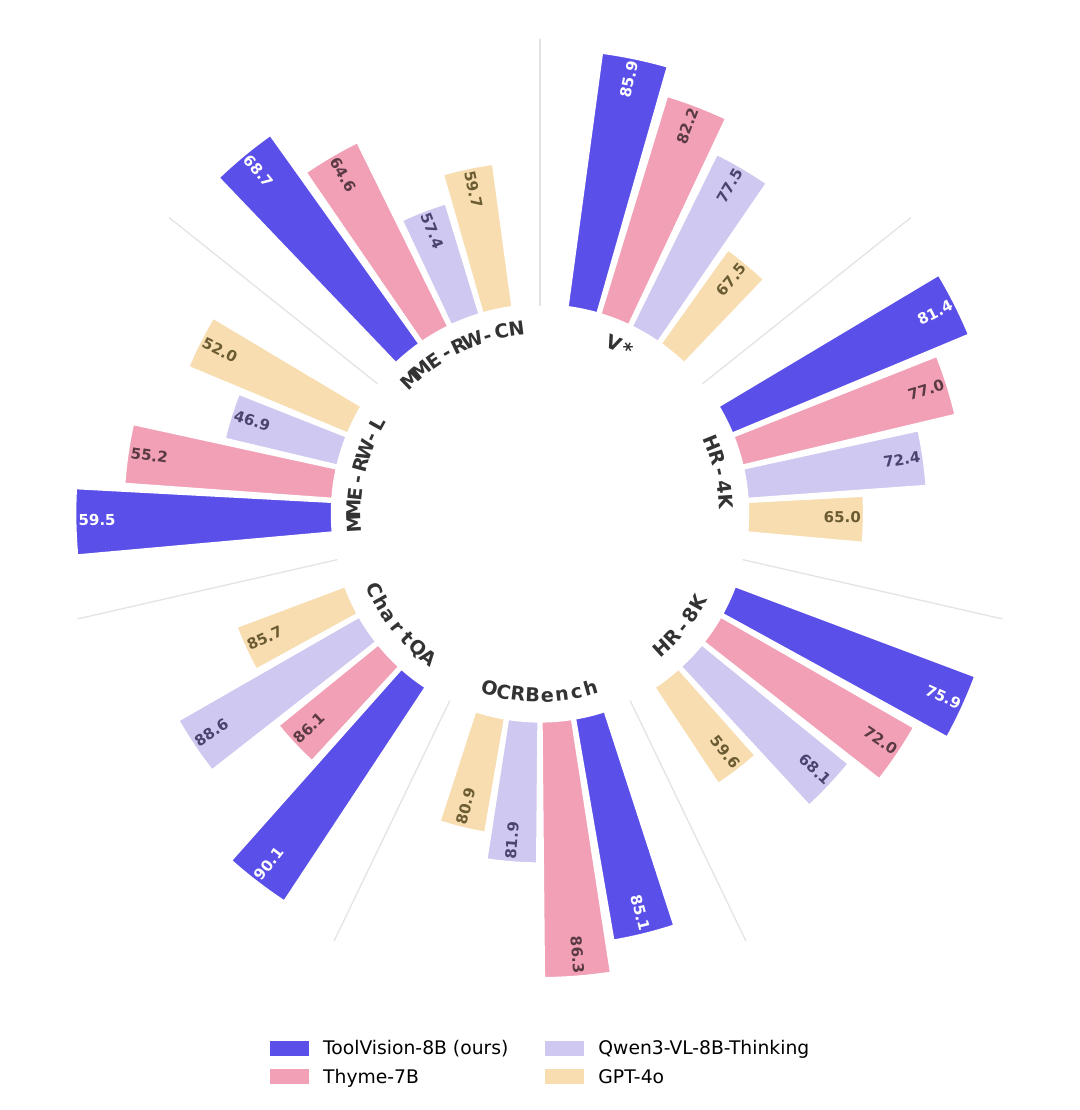}
\caption{ToolVision-8B against its base model (Qwen3-VL-8B-Thinking),
Thyme-7B, and GPT-4o across the seven main benchmarks. Bar lengths are
scaled within each benchmark; labels give absolute scores.}
\label{fig:teaser}
\end{figure}

\section{Introduction}

\begin{figure*}[t]
\centering
\includegraphics[width=0.97\textwidth]{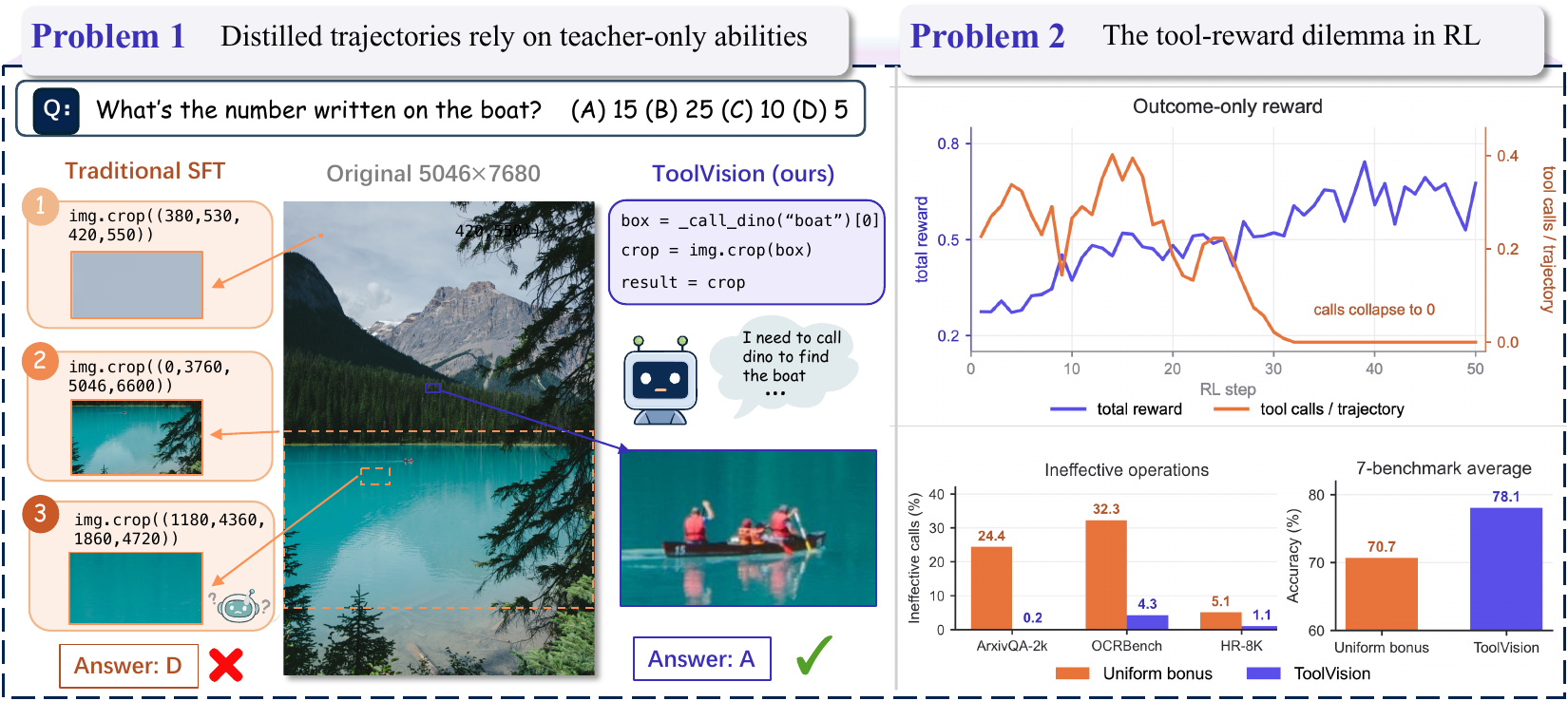}
\caption{\textbf{Two failure modes in the SFT-then-RL recipe.} \textbf{Left}:
coordinate-crop failures in the distilled-then-RL policy, alongside
detector-guided localization in ToolVision.
\textbf{Right, top}: total reward and tool calls per trajectory under
outcome-only optimization. \textbf{Right, bottom}: ineffective-operation
rates and the seven-benchmark average for the uniform-bonus variant
($w\equiv0.2$ for every question) and ToolVision. Ineffective operations are
unmodified-image re-displays or unnecessary rotate/flips.}
\label{fig:problems}
\end{figure*}

Multimodal large language models (MLLMs) can now reason about an image by
acting on it rather than describing it once. Under the
\emph{thinking with images} paradigm popularized by o3~\citep{openai2025o3},
a model interleaves reasoning with tool calls that transform or query the
visual input, recovering evidence a single forward pass would miss
~\citep{bai2025qwenvl,chen2024internvl}. Recent systems replace early fixed
crop-and-zoom operations~\citep{wu2024vstar,wang2025hrbench} with a
\emph{code-as-tool} interface, letting the model write executable code
to invoke arbitrary image operations~\citep{guo2026codevision,zhang2026thyme}.
The promise is a form of leverage. A model may zoom into a distant region,
read dense text with an OCR tool~\citep{cui2025paddleocr}, or delegate
localization to an open-vocabulary detector~\citep{liu2023groundingdino};
by routing perception through the right tool, even a modestly sized model
can reach answers that would otherwise demand far stronger perception. Yet for an 8B
learner, the prevailing SFT-then-RL recipe breaks down in a different way at
each stage (\textbf{Figure~\ref{fig:problems}}).

\noindent\textbf{Misalignment in How to Use Visual Tools.}
Existing work commonly uses SFT to distill tool-use trajectories from a much
stronger teacher~\citep{zhang2026thyme}. The teacher chooses tools according to
its own visual capabilities. For example, it may read small text directly when
the student needs an OCR tool, or predict crop coordinates from fine-grained
visual cues that the student cannot reliably perceive. Traditional SFT
imitates the complete teacher trajectory without checking whether the student
can reproduce the capabilities behind it, so the student copies tool-call
patterns without learning to obtain useful visual evidence. As shown in
\textbf{Figure~\ref{fig:problems}} (left), the distilled policy does invoke the
crop tool, but it follows the teacher's direct coordinate-prediction pattern
and misses the target because it lacks the same localization ability. The
problem is therefore not that the teacher trajectory is invalid, but that the
demonstration is misaligned with the student's capabilities.\looseness=-1

\noindent\textbf{Misalignment in When to Use Visual Tools.}
RL supervision faces two opposite failures, neither of which teaches the model
when tool use is worthwhile given its own capabilities. Without a tool-specific
incentive, each tool call is a liability because code generation, execution,
or perception may fail. Under outcome-only optimization, avoiding tools can
therefore be safer than exploring them. As shown in \textbf{Figure~\ref{fig:problems}}
(right, top), total reward continues to rise while tool calls fall to zero,
consistent with Thyme's observation that unreliable code generation
pressures the policy toward direct answers~\citep{zhang2026thyme}. At the opposite
extreme, giving the same bonus to every correct tool-using trajectory teaches
the policy that tool use is generally profitable, even when it provides no
useful evidence. As shown in \textbf{Figure~\ref{fig:problems}} (right, bottom), this
produces more ineffective operations while reducing average accuracy. The RL
problem is therefore not simply to preserve or increase tool use, but to reward
it only when it provides useful evidence for the learner on that question.\looseness=-1

\begin{figure*}[t]
\centering
\includegraphics[width=0.98\textwidth]{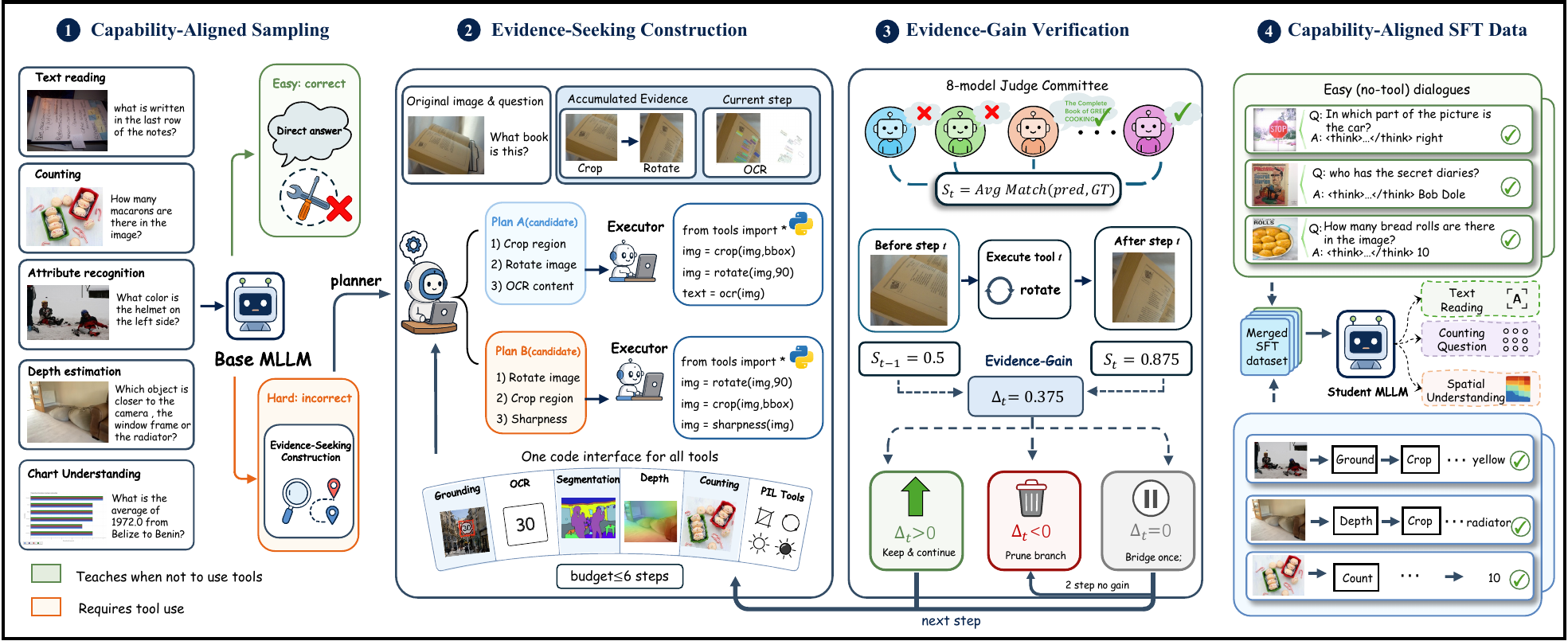}
\caption{\textbf{Overview of the capability-aligned SFT construction pipeline.}
Base-model successes provide direct-answer dialogues, while failures enter
Planner--Executor search over the shared visual-tool interface. An eight-model
committee scores stepwise evidence gain to retain or prune branches; validated
trajectories merge with direct dialogues for SFT.}
\label{fig:pipeline}
\end{figure*}

To address these two misalignments, we introduce ToolVision, which aligns
supervision for how and when to use visual tools with the learner's
capabilities. For SFT, a multi-agent pipeline proposes and executes candidate
tool operations through a unified code interface, and a cross-scale committee
scores the evidence each step yields for student-scale models; only verified
trajectories enter SFT. For RL, ToolVision compares the frozen SFT model's
performance with and without tools on each question, and grants the
tool-success reward only where tools provide a clear benefit. Both stages
share a heterogeneous visual toolbelt spanning OCR, open-vocabulary detection
and segmentation, depth estimation, counting, and pure image operations
(Section~\ref{sec:prelim}). \textbf{Figure~\ref{fig:pipeline}} summarizes the
pipeline, with details in Section~\ref{sec:toolvision_sft}.

ToolVision-8B improves over its base on all seven main benchmarks
(\textbf{Figure~\ref{fig:teaser}}). It reaches $85.9$ on V\textsuperscript{*}, $81.4$ on
HRBench 4K, and $75.9$ on HRBench 8K, surpassing Thyme-7B, CodeVision-8B, and
CodeDance-7B on all three. Despite using only 8B parameters, it
outperforms Qwen3-VL-32B-Thinking on V\textsuperscript{*} and HRBench 8K.
Behavioral analysis shows that specialist operations are used more
often on tasks where their evidence is relevant, including text
reading, counting, and small-target localization.

In summary, our contributions are threefold:
\begin{itemize}
  \item We identify capability misalignment in both how visual-tool use is
  demonstrated during SFT and when it is rewarded during RL. ToolVision aligns
  both stages using supervision built automatically from public task data.\looseness=-1
  \item For SFT, we introduce capability-aligned trajectory synthesis, which
  retains only trajectories whose stepwise evidence measurably helps
  student-scale models, teaching executable tool behaviors rather than
  surface-level tool-call patterns.\looseness=-1
  \item For RL, we introduce model-conditioned must-use-tool (MUT)
  supervision, whose fixed necessity labels gate the tool-success reward only
  where tools demonstrably help the frozen SFT model, avoiding both tool
  collapse and question-agnostic overuse.
\end{itemize}

\begin{table*}[t]
\centering
\small
\setlength{\tabcolsep}{8pt}
\begin{tabular}{l ccc cc cc}
\toprule
 & \multicolumn{3}{c}{High-Resolution} & \multicolumn{2}{c}{Text \& Chart} & \multicolumn{2}{c}{Real World} \\
\cmidrule(lr){2-4} \cmidrule(lr){5-6} \cmidrule(lr){7-8}
Model & V\textsuperscript{*} & HR-4K & HR-8K & OCRBench & ChartQA & MME-L & MME-CN \\
\midrule
\multicolumn{8}{c}{\textit{Closed-Source MLLMs}} \\
GPT-4o$^\dag$ & 67.5 & 65.0 & 59.6 & 80.9 & 85.7 & 52.0 & 59.7 \\
\midrule
\multicolumn{8}{c}{\textit{Open-Source MLLMs}} \\
Qwen2.5-VL-7B$^\dag$ & 76.4 & 68.8 & 65.3 & \textbf{88.4} & 83.7 & 44.1 & 60.8 \\
Qwen2.5-VL-32B$^\dag$ & 81.2 & 73.4 & 70.4 & 85.5 & 81.1 & 46.2 & 60.5 \\
InternVL3-8B$^\dag$ & 70.2 & 70.0 & 69.3 & 88.1 & 85.9 & 48.6 & 60.5 \\
Qwen3-VL-30B-A3B-Thinking$^\ddag$ & 81.2 & 77.8 & 71.3 & 83.9$^\S$ & --- & --- & --- \\
Qwen3-VL-32B-Thinking$^\ddag$ & 84.8 & \textbf{82.1} & 74.8 & 85.5$^\S$ & --- & --- & --- \\
Qwen3-VL-8B-Thinking (base) & 77.5 & 72.4 & 68.1 & 81.9 & 88.6 & 46.9 & 57.4 \\
\midrule
\multicolumn{8}{c}{\textit{Tool-Integrated Open-Source MLLMs}} \\
Thyme-7B$^\dag$ & 82.2 & 77.0 & 72.0 & 86.3 & 86.1 & 55.2 & 64.6 \\
CodeVision-8B$^\ddag$ & 82.4 & 77.1 & 73.4 & --- & --- & --- & --- \\
CodeDance-7B$^\clubsuit$ & 84.8 & 75.2 & 72.3 & --- & 87.5 & --- & --- \\
\textbf{ToolVision-8B (ours)} & \textbf{85.9}\rlap{\,{\scriptsize(+8.4)}} & 81.4\rlap{\,{\scriptsize(+9.0)}} & \textbf{75.9}\rlap{\,{\scriptsize(+7.8)}} & 85.1\rlap{\,{\scriptsize(+3.2)}} & \textbf{90.1}\rlap{\,{\scriptsize(+1.5)}} & \textbf{59.5}\rlap{\,{\scriptsize(+12.7)}} & \textbf{68.7}\rlap{\,{\scriptsize(+11.3)}} \\
\bottomrule
\end{tabular}
\caption{\textbf{Main results on seven benchmarks.} Sources: $\dag$ Thyme~\citep{zhang2026thyme};
$\ddag$ CodeVision~\citep{guo2026codevision}; $\clubsuit$ CodeDance~\citep{song2025codedance};
and $\S$ the Qwen3-VL technical report~\citep{bai2025qwen3vl}.
``---'' denotes unreported results. MME-L/CN denote
MME-RealWorld-Lite and MME-RealWorld-CN. Bold marks the column best;
parenthesized deltas are improvements over the base model, computed before
rounding the displayed scores.}
\label{tab:main}
\end{table*}

\begin{table}[t]
\centering
\small
\begin{tabular}{@{}l@{\hspace{10pt}}c@{\hspace{10pt}}c@{\hspace{2.5em}}}
\toprule
Model & ArxivQA-2k $\uparrow$ & FSC-147 $\downarrow$ \\
\midrule
Qwen3-VL-8B-Thinking & 58.3 & 44.70 \\
\textbf{ToolVision-8B (ours)} & \textbf{74.5}\rlap{\,{\scriptsize(+16.2)}} & \textbf{11.56}\rlap{\,{\scriptsize($-74\%$)}} \\
\bottomrule
\end{tabular}
\caption{\textbf{Results on ArxivQA-2k and FSC-147-test.} Parentheses report the
accuracy gain and relative MAE reduction over the base model, respectively.}
\label{tab:tool_heavy}
\end{table}

\section{Related Work}

\paragraph{Visual Tool Interfaces.}
Systems that implement \emph{thinking with images}~\citep{openai2025o3}
differ mainly in their tool interfaces. Early systems bind this ability to a small set of
fixed operations---guided visual search, crop-and-zoom, region-level
highlighting---that augment but do not transcend the model's original visual
encoder~\citep{wu2024vstar,zheng2025deepeyes,wang2025pixelreasoner,
liu2024chainofspot,shao2024visualcot,fan2025grit,wu2025vilasr,yang2023som}.
A complementary line of visual-programming methods compiles a natural-language
query into a program over vision APIs, obtaining compositional reasoning without
training~\citep{suris2023vipergpt,gupta2023visprog,hu2024sketchpad}. More
recent work replaces the fixed tool registry with a \emph{code-as-tool}
interface in which the model writes executable code to invoke arbitrary image
operations~\citep{zhao2025pyvision,zhang2026thyme,song2025codedance,
guo2026codevision}; see \citet{su2025twisurvey} for a survey. These systems make
diverse visual operations executable; the remaining question is how to
supervise which operations produce usable evidence and when they are worth
invoking.\looseness=-1

\paragraph{Learning Visual Tool Use.}
Group-relative policy optimization~\citep{shao2024grpo,yu2025dapo,
zheng2025gspo,guo2025deepseekr1} has become a standard approach for eliciting
reasoning, and a parallel line integrates external tools into the reasoning
loop~\citep{yao2023react,schick2023toolformer,gou2024tora,feng2025retool,
qian2025toolrl,jin2025searchr1,liu2025visualrft}. In visual tool use,
outcome-only objectives and blanket tool bonuses reproduce the two extremes
discussed in the introduction~\citep{zheng2025deepeyes,wang2025pixelreasoner}. Among code-as-tool
systems, Thyme~\citep{zhang2026thyme} activates broad code capabilities through
large-scale curated SFT and uses an outcome-centered RL objective augmented
with format and reasoning-consistency rewards.
CodeVision~\citep{guo2026codevision} strengthens this recipe with dense process
supervision constructed from required-tool metadata, image transformations,
and ground-truth crop regions. This guidance makes the target operation
explicit, but also predefines the tool distribution and requires new rules and
metadata as tasks and tools expand. CodeDance~\citep{song2025codedance} takes a
different route: rather than specifying a required operation, it adapts tool
incentives using the aggregate accuracy of the current rollout group and
execution feedback. Empirical difficulty, however, is not the counterfactual
benefit of tools: a low-accuracy group may represent either a tool-responsive
question or one that neither the model nor its tools can solve; successful
execution need not yield useful evidence; and the online proxy changes with the
policy. ToolVision instead verifies evidence gain across model scales for SFT
and measures tool necessity offline for the frozen SFT model.

\section{Methodology}

\subsection{Preliminaries}\label{sec:prelim}

\subsubsection{Problem Setup and Supervision Targets.}
Let $x=(I,q)$ denote an image--question pair with reference answer $y$. An MLLM
policy $\pi_\theta$ solves $x$ over multiple turns. At turn $t$, it either emits a
final answer or produces a code action $a_t$ over the exposed operations
$\mathcal{O}$. Executing $a_t$ in a sandbox returns an image or text result
$z_t$, which is appended to the accumulated evidence
$E_t=(I,z_1,\ldots,z_t)$. A tool step is useful only when the returned result
becomes evidence that the learner can exploit.\looseness=-1

We formalize this requirement with the cross-scale Judge committee
$\mathcal{C}$ introduced below. Its answerability score after step $t$ is
\[
\begin{aligned}
J(E_t;x)
  &=\frac{1}{|\mathcal{C}|}\sum_{m\in\mathcal{C}}
    \operatorname{score}\!\left(m(q,E_t),y\right),\\
\Delta_t
  &=J(E_t;x)-J(E_{t-1};x).
\end{aligned}
\]
Here $\operatorname{score}(\cdot)$ grades each member's answer to $q$, produced
from the evidence $E_t$, against the reference $y$.
The stepwise difference $\Delta_t$ is a capability-sensitive evidence-gain
signal: it measures whether the new result improves answerability across model
scales, rather than whether a strong proposer can use it by itself.

Tool necessity is a separate, question-level target. We write
$w(x,\pi_{\mathrm{SFT}})$ for the necessity confidence obtained by comparing
paired rollouts of the frozen SFT policy $\pi_{\mathrm{SFT}}$ with tools
disabled and enabled. Its construction is specified in
Section~\ref{sec:mut_measurement}. Making the
model explicit is
essential: the same question may require a tool for one policy but not for a
stronger one.
Together, the two quantities align supervision at complementary stages:
$\Delta_t$ guides SFT toward evidence usable at the learner's scale, whereas
$w(x,\pi_{\mathrm{SFT}})$ guides RL toward questions that benefit from tools.
SFT thereby places useful behaviors within the policy's \emph{practical
support}---making them likely enough to be explored---and RL selectively
reweights and amplifies them.

\subsubsection{A Heterogeneous Visual Toolbelt.}
The two supervision targets become meaningful only when the interface exposes
heterogeneous sources of visual evidence rather than geometric edits alone.
Behind a single code interface, we assemble a toolbelt that spans specialist
services and a set of pure image operations. An optical character recognition
(OCR) tool reads dense text~\citep{cui2025paddleocr}; an open-vocabulary
detector paired with a promptable segmenter localizes, crops, and masks objects
by text phrase~\citep{liu2023groundingdino,ravi2024sam2}; a monocular estimator
returns metric depth~\citep{bochkovskii2025depthpro}; and an open-world counter
returns instance counts~\citep{amininaieni2024countgd}. Alongside these, pure
operations---cropping, resizing, rotation, flipping, contrast and sharpness
adjustment, and drawing---handle geometric and photometric transforms without
external models. This heterogeneity makes both decisions nontrivial: whether a
returned result becomes usable evidence for student-scale models, and whether
a given question--model pair benefits from invoking a tool. All specialist
services fit within 16~GB of GPU memory on a single shared GPU. Deployment
details are provided in the Appendix and the accompanying code release.\looseness=-1

\subsection{Capability-Aligned Supervised Fine-Tuning}\label{sec:toolvision_sft}
The supervised stage addresses the teacher--student capability misalignment by
constructing tool trajectories primarily from questions the base model fails,
while retaining a smaller set of short direct-answer trajectories to preserve
the option of answering without a tool (\textbf{Figure~\ref{fig:pipeline}}).\looseness=-1

\subsubsection{SFT Dataset.}
Our SFT set contains $4{,}057$ examples drawn from public training splits that
span compositional and spatial reasoning, text reading, counting, and
orientation correction~\citep{hudson2019gqa,daxberger2025mmspatial,
singh2019textvqa,ranjan2021fsc147,guo2026codevision}. CodeVision training
examples are used only to provide seed demonstrations for the reversible
rotate and flip operations. All
remaining tool calls and step-level values are synthesized without additional
human annotations of tool use or step utility.

\subsubsection{Capability-Aligned Trajectory Synthesis.}
For each base-model failure, a Planner observes the accumulated evidence and
either proposes up to two alternative next actions or decides that the question
can be answered. An Executor realizes each proposal as Python code over the
toolbelt and runs it in the sandbox. Both roles use Qwen3.6-plus, but a strong
model is used only to propose and execute candidates: an action does not become
supervision merely because the teacher produced it. The search retains at most
two child trajectories after each round and allows at most six executed steps
along any trajectory.\looseness=-1

Execution provides the first filter. Exceptions while producing the code or
running it in the sandbox mark the branch as erroneous; an unsuccessful tool
return marks it as failed; and a branch whose return contains neither an image
nor text is pruned. Such branches leave the search frontier and cannot enter the exported
SFT data.

Every successfully executed step is evaluated by an eight-model committee
spanning strong multimodal models, student-scale models, and multiple model
families. The full roster, scoring protocol, and a proposer-sensitivity check
are given in the Appendix. Strong members stabilize the estimate
of whether the accumulated evidence remains sufficient, while student-scale
members keep the score sensitive to evidence that becomes usable at the
target scale. The committee score ranks the search frontier and controls
termination: a branch is pruned immediately when its score decreases, a
single zero-gain step is tolerated, and a second consecutive zero-gain step
terminates the branch. Completed trajectories are finally checked for a
correct final answer, and only those are exported as supervised examples.
The resulting examples seed executable tool behaviors whose returned evidence
is useful beyond the strong model that proposed them.

The resulting checkpoint, referred to below as the frozen SFT model,
initializes RL.

\subsection{Necessity-Aware Reinforcement Learning}
\label{sec:mut_measurement}
Reinforcement learning can in principle discover new behavior, but in practice
it primarily sharpens the practical support seeded by supervised fine-tuning.
We optimize the policy with Group Sequence Policy Optimization
(GSPO)~\citep{zheng2025gspo}. Tool incentives face the two extremes identified
in the introduction: outcome-only rewards suppress tool exploration, while
question-agnostic bonuses encourage valid but ineffective operations. A useful
tool incentive must therefore depend on both the question and the model being
trained.

\subsubsection{Measuring Tool Necessity and Constructing the RL Set.}
As noted above, tool necessity is model-dependent: the same question may
require an OCR or localization tool for the frozen SFT model while a stronger
model can answer it directly. Difficulty alone is also insufficient, because a question
that is difficult without tools may either become solvable with tools or remain
beyond both the model and its tools. We therefore estimate
$w(x,\pi_{\mathrm{SFT}})$ once before RL using a sequential rollout procedure
and keep the resulting confidence fixed throughout training. This contrast
also addresses a blind spot of conventional outcome filtering: all-correct and
all-wrong rollout groups have zero group-relative advantage, yet some questions
in the latter group become solvable with tools.\looseness=-1

We first sample sixteen responses from the frozen SFT model with tools
disabled. Questions answered correctly at most eight times are retained as
tool-benefit candidates. This stage identifies questions on which the model is
unreliable, without assuming that a tool can actually help. For each candidate,
we then expose the full tool interface and sample eight additional agent
trajectories. Tool use is optional: the model may invoke one or more tools or
answer directly. We define the grounded tool-correct count (\emph{GTC}) as the number of correct
trajectories in the tool-enabled arm that choose to invoke at least one tool.
The no-tool correct count (\emph{NTC}) records correct trajectories that
answer directly without invoking a tool in this tool-enabled stage.
Since this measurement precedes RL, no reward yet favors tool calls, so a
policy that can answer from internal knowledge simply answers directly---and
such successes are precisely what the no-tool rollouts and NTC record. A
question enters the necessity-weighted tiers only when correct answers
co-occur with tool use and rarely arise without it, so the GTC--NTC contrast
acts as a question-level counterfactual signal, not a per-trajectory
usage check.

We use these counts to form three training groups. Strong MUT questions show
repeated correct tool-using trajectories but little direct success
($\mathrm{GTC}\geq2$, $\mathrm{NTC}\leq1$) and receive $w=0.5$. Questions with a
single correct tool-using trajectory ($\mathrm{GTC}=1$) form the weak group and
receive $w=0.2$. To preserve direct reasoning, we then sample ordinary
questions---stratified by source---from those answered correctly in nine to
fifteen of the sixteen no-tool rollouts and set $w=0$. Questions always solved without tools
and hard candidates without clear tool benefit are excluded.

Beyond the SFT sources, the RL pool uses publicly released splits spanning chart
understanding, scientific figures, spatial reasoning, and counting, including
ChartQA, ArxivQA, and
PixMo-Count~\citep{masry2022chartqa,li2024arxivqa,deitke2024molmo}. The full
source list, split choices, and per-source counts are provided in the Appendix. The final set contains $24{,}921$ questions: $8{,}460$ strong MUT,
$6{,}481$ weak, and $9{,}980$ ordinary.

\subsubsection{Necessity-Conditioned Reward.}
Writing $n$ for the number of tool calls, we optimize
\[
R = R_\text{acc} + \lambda\,R_\text{proto}
      + w(x,\pi_{\mathrm{SFT}})\,R_\text{mut}
      - \mu\,\max(0,\, n - \tau),
\]
where $R_\text{acc}$ marks a correct final answer. $R_\text{proto}$ is a
format-and-protocol gate: the reasoning and answer tags must be well formed and,
when a tool call is present, its JSON must be valid. It is independent of
whether a tool is used, so a valid direct answer can receive the full protocol
score. The trajectory-level indicator $R_\text{mut}\in\{0,1\}$ marks tool-use
success: the final answer is correct, at least one tool is used, and no tool
call raises an execution error. The frozen label confidence
$w(x,\pi_{\mathrm{SFT}})$ determines whether, and how strongly, a successful
tool-using trajectory is rewarded through the product
$w(x,\pi_{\mathrm{SFT}})R_\text{mut}$.

The final term penalizes calls beyond an overuse threshold $\tau$. Ordinary questions have
$w=0$ but retain the accuracy, protocol, and overuse terms, so nothing in their
reward favors calling a tool. On a necessity-weighted question answered without
a tool, $R_\text{mut}=0$ but the answer is not penalized. The reward therefore
selectively amplifies successful tool behavior where the paired measurement
finds it useful, rather than rewarding tool use on every question.

\section{Experiments}

\begin{figure*}[!t]
\centering
\includegraphics[width=0.94\textwidth]{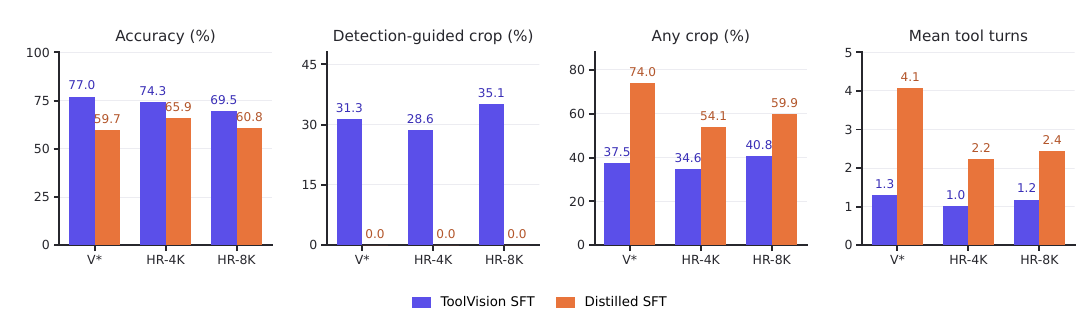}
\caption{\textbf{Pre-RL comparison of Distilled SFT and ToolVision SFT on three
high-resolution benchmarks} (left to right: accuracy, detection-guided crop
rate, any-crop rate, mean tool turns). Distilled SFT crops more images and
takes more tool turns, yet never invokes detection-guided cropping and scores
lower: it inherits the teacher's coordinate-crop pattern without the
perception to aim it. ToolVision SFT delegates localization to the detector
and answers in fewer turns.}
\label{fig:dist}
\end{figure*}

\begin{table*}[!t]
\centering
\small
\setlength{\tabcolsep}{2.0pt}
\begin{tabular}{l ccc cc cc cc}
\toprule
& \multicolumn{3}{c}{High resolution} & \multicolumn{2}{c}{Reading}
& \multicolumn{2}{c}{Real world} & \multicolumn{2}{c}{Specialized} \\
\cmidrule(lr){2-4}\cmidrule(lr){5-6}\cmidrule(lr){7-8}\cmidrule(lr){9-10}
Model & V\textsuperscript{*} & HR-4K & HR-8K & OCRBench & ChartQA
& MME-L & MME-CN & ArxivQA-2k & FSC-147 $\downarrow$ \\
\midrule
\multicolumn{10}{c}{\textit{Base model}} \\
Qwen3-VL-8B-Thinking & 77.5 & 72.4 & 68.1 & 81.9 & 88.6 & 46.9 & 57.4 & 58.3 & 44.70 \\
\midrule
\multicolumn{10}{c}{\textit{After SFT}} \\
Distilled SFT & 59.7 & 65.9 & 60.8 & \textbf{78.2} & 84.8 & 41.8 & 42.9 & 59.5 & 16.11 \\
ToolVision SFT & \textbf{77.0} & \textbf{74.3} & \textbf{69.5} & \textbf{78.2}
& \textbf{87.8} & \textbf{52.7} & \textbf{54.4} & \textbf{66.9} & \textbf{15.77} \\
\midrule
\multicolumn{10}{c}{\textit{After RL}} \\
Distilled SFT + MUT RL & 83.8 & 78.9 & 75.3 & 83.3
& \textbf{90.2} & 58.6 & 68.2 & 71.8 & 13.53 \\
Uniform bonus ($w\equiv0.2$) & 77.0 & 76.0 & 68.6 & 81.7 & 88.4 & 52.3 & 50.6 & 69.1 & 12.16 \\
ToolVision-8B & \textbf{85.9} & \textbf{81.4} & \textbf{75.9} & \textbf{85.1}
& 90.1 & \textbf{59.5} & \textbf{68.7} & \textbf{74.5} & \textbf{11.56} \\
\bottomrule
\end{tabular}
\caption{\textbf{Stage-wise results on nine benchmarks.} Bold: best within each
block; FSC-147 reports MAE (lower is better).}
\label{tab:stage_ablation}
\end{table*}

\subsection{Experimental Settings}
ToolVision-8B is initialized from Qwen3-VL-8B-Thinking and fine-tuned on the
$4{,}057$ SFT examples described above. We train the language model for two
epochs with a global batch size of $128$, freezing the vision encoder and
projector. Optimizer settings are listed in the Appendix.
Reinforcement learning then runs GSPO on the $24{,}921$-question training
mixture with 128 prompts per batch, eight rollouts per prompt, a
mini-batch size of 32, and an actor learning rate of $10^{-6}$. Rollouts use a
maximum of 12 assistant turns, temperature $0.7$, and top-$p$ $0.95$. We set $\lambda=0.2$,
$\mu=0.05$, the tool-call overuse threshold $\tau=6$, and confidence weights $w=0.5/0.2/0$
for strong MUT, weak, and ordinary questions. We report the
checkpoint at step $180$. Benchmark instances and metric implementations are
aligned with LMMs-Eval~\citep{zhang2024lmmseval}. Metrics follow each
benchmark's standard protocol (multiple-choice accuracy; ChartQA relaxed
accuracy; OCRBench inclusion-based score; FSC-147 MAE, lower is better), with
details in the Appendix.
ToolVision rollouts use our multi-turn agent harness, after which final answers
are scored with the same benchmark-specific logic. Following CodeVision's
released evaluation procedure~\citep{guo2026codevision}, we use a fixed,
zero-temperature, reference-conditioned LLM judge only when the benchmark's
rule-based matcher cannot reliably resolve semantic equivalence.
At evaluation, we sample four trajectories per question at temperature $0.7$
and report avg@4 as the primary accuracy-oriented metric. Across the four
evaluation runs, the per-benchmark standard deviation of ToolVision-8B ranges
from $0.17$ (MME-RealWorld-CN) to $0.99$ (HRBench 4K); the Appendix reports the full table.
We will release the training and evaluation code, data-construction scripts,
and configurations.\looseness=-1

\subsection{Benchmarks and Baselines}
We evaluate performance on nine benchmarks covering high-resolution perception
(V\textsuperscript{*}~\citep{wu2024vstar}, HRBench 4K/8K~\citep{wang2025hrbench}),
text and chart reading (OCRBench~\citep{liu2024ocrbench},
ChartQA~\citep{masry2022chartqa}), real-world scenes
(MME-RealWorld-Lite/CN~\citep{zhang2024mmerealworld}), scientific figures
(ArxivQA-2k~\citep{li2024arxivqa}), and counting
(FSC-147-test~\citep{ranjan2021fsc147}). \textbf{Table~\ref{tab:main}} reports the
seven benchmarks for which external baselines have publicly reported numbers.
Before training-data preprocessing, we randomly sample and fix $2{,}000$
ArxivQA questions for evaluation and exclude their question identifiers from
the training mixtures. This ArxivQA-2k subset and FSC-147-test appear in
\textbf{Table~\ref{tab:tool_heavy}}.

Baselines fall into four groups: the closed-source GPT-4o; open-source MLLM
baselines Qwen2.5-VL-7B/32B and InternVL3-8B~\citep{zhu2025internvl3}; larger
reasoning models
Qwen3-VL-30B-A3B-Thinking and Qwen3-VL-32B-Thinking; and the three closest
code-as-tool systems, Thyme-7B, CodeVision-8B, and CodeDance-7B (built on
Qwen2.5-VL-7B), together with our base model. Source symbols follow
\textbf{Table~\ref{tab:main}}; unreported entries are marked ``---''.

\subsection{Main Results}
Across all seven benchmarks in \textbf{Table~\ref{tab:main}}, ToolVision-8B improves
over its base. The largest gains appear on MME-RealWorld-Lite (\textbf{+12.7}),
MME-RealWorld-CN (\textbf{+11.3}), and HRBench 4K (\textbf{+9.0}).\looseness=-1

Against the three closest code-as-tool systems, ToolVision-8B leads on all
three high-resolution benchmarks (\textbf{Table~\ref{tab:main}}). Since part of the
absolute margin reflects base-model differences, we compare each system's
gain over its own base: our gains match or exceed all three on every
high-resolution benchmark, reaching $+9.0$ on HRBench 4K.
CodeVision-8B builds on the same Qwen3-VL-8B-Thinking as ToolVision, so its
absolute gap (up to $+4.3$) reflects the method rather than the base.
The comparison with larger reasoning models points the same way:
ToolVision-8B exceeds Qwen3-VL-30B-A3B-Thinking on every benchmark both
report, and, at a quarter of the parameter count, surpasses
Qwen3-VL-32B-Thinking on both V\textsuperscript{*} and HRBench 8K. On
perception-heavy tasks, access to useful specialist evidence can partly
offset model scale.

The gains concentrate on tool-dependent tasks (\textbf{Table~\ref{tab:tool_heavy}}):
ArxivQA-2k rises from $58.3$ to $74.5$ (\textbf{+16.2}), and the FSC-147-test
MAE drops from $44.7$ to $11.6$, a \textbf{74\%} reduction.

\subsection{Stage-wise Ablations and Analysis}

\textbf{Table~\ref{tab:stage_ablation}} organizes the analysis by training stage: the
base model, the two supervised initializations, and two component ablations
alongside the complete ToolVision-8B model.

\subsubsection{Capability-Aligned SFT.}
Distilled SFT distills trajectories from Qwen3.6-plus on the same question
pool, with the same toolbelt and code interface. Before RL,
ToolVision SFT outperforms Distilled SFT on eight of the nine
benchmarks and matches it on OCRBench. The largest gains are \textbf{+17.3} on
V\textsuperscript{*}, \textbf{+11.5} on MME-RealWorld-CN, and \textbf{+10.9} on
MME-RealWorld-Lite.

\textbf{Figure~\ref{fig:dist}} examines the three high-resolution benchmarks before RL.
Distilled SFT trails ToolVision SFT by $8$--$17$ accuracy points, crops on
$54$--$74\%$ of questions versus our $35$--$41\%$, and spends $2$--$3\times$
more tool turns. Crucially, none of its crops are detection-guided: the
distilled policy inherits the teacher's habit of emitting raw crop
coordinates, but not the fine-grained perception that made those coordinates
accurate. On questions beyond its own perception, it still predicts
coordinates rather than delegating localization to the detector. ToolVision
SFT delegates on $29$--$35\%$ of questions and answers in fewer turns. Thus
the SFT gain comes not from more tool activity, but from localization
evidence that the student can use reliably.\looseness=-1

\subsubsection{Stage-wise Component Ablations.}
To isolate the SFT stage, we apply the same MUT RL training to Distilled SFT.
The resulting Distilled SFT + MUT RL model scores below ToolVision-8B on seven
of the eight accuracy benchmarks and has a higher FSC-147 MAE. This shows that MUT RL does not remove the need for capability-aligned
SFT. The score gap is accompanied by a behavioral difference. In a separate
behavior analysis on the same three benchmarks, the Distilled SFT + MUT RL model uses
detection-guided cropping on only $0.0\%$, $0.5\%$, and $0.1\%$ of questions,
respectively; \textbf{Figure~\ref{fig:problems}} (left) illustrates a coordinate-crop
failure.\looseness=-1

The reward design must also avoid two opposite failures. Without a tool-success
reward, tool calls collapse to zero even as total reward rises
(\textbf{Figure~\ref{fig:problems}}, right top). At the other extreme, the uniform-bonus
variant fixes $w=0.2$ for every question. It retains $R_\text{mut}$, so every
correct, execution-error-free trajectory that uses at least one tool receives
the same bonus, regardless of whether the question benefits from tool use.
ToolVision instead uses necessity-conditioned weights of $0.5$, $0.2$, and
$0$: it increases the reward for strong MUT questions, keeps the weak tier
unchanged, and removes the tool bonus from ordinary questions. The
uniform-bonus variant underperforms ToolVision-8B on all nine benchmarks, by up
to $18.1$ points on MME-RealWorld-CN, and produces more ineffective operations
(\textbf{Figure~\ref{fig:problems}}, right bottom). These results show that the benefit
comes from conditioning the tool-success reward on measured necessity rather
than from rewarding successful tool use indiscriminately.\looseness=-1

After RL, tool choice specializes by task: ToolVision-8B invokes the counting
tool on $100\%$ of FSC-147 questions, OCR on $84\%$ of OCRBench questions, and
open-vocabulary grounding on $72\%$ of V\textsuperscript{*} questions, while
manual coordinate cropping all but disappears ($\leq0.1\%$ on every
benchmark). The Appendix reports the detailed breakdown.\looseness=-1

\section{Conclusion}
We study why the standard SFT-then-RL recipe fails to teach smaller multimodal
models to use visual tools effectively. ToolVision addresses the two
misalignments with capability-aligned trajectory synthesis and
necessity-gated tool rewards. ToolVision-8B
improves over its base across all seven main benchmarks and outperforms the closest
code-as-tool systems on all three high-resolution benchmarks, with substantial
gains on scientific-figure understanding and counting.\looseness=-1

{\small
\bibliography{references}}

\appendix

\section{Detailed Algorithms}

The two procedures below expand the supervised trajectory construction and MUT
measurement described in the main text.

\subsection{Capability-Aligned SFT Trajectory Synthesis}

The procedure separates proposal, execution, and verification. A strong model
may propose and execute an operation, but the operation becomes a training
target only after successful execution and cross-scale evidence-gain
verification.

\begin{figure*}[!t]
\small
\hrule
\vspace{0.5em}
\noindent\textbf{Algorithm 1: Capability-aligned SFT trajectory synthesis}
\begin{algorithmic}[1]
\REQUIRE Question $x$, answer $y$, toolbelt $\mathcal{O}$, committee
$\mathcal{C}$; depth $D=6$, Planner proposal budget $B=2$, beam width $K=2$
\STATE Initialize the frontier with original evidence $E_0=(I)$
\FOR{$t=1,\ldots,D$ and while the frontier is nonempty}
  \STATE Ask the Planner for candidate set $\mathcal{A}_t$ with
  $|\mathcal{A}_t|\leq B$ for each frontier state
  \FOR{each $a\in\mathcal{A}_t$}
    \STATE Execute $a$ over $\mathcal{O}$; discard errors and invalid returns
    \STATE Append the valid return to the evidence $E_t$ and compute the
    committee score $J(E_t;x)$
    \STATE $\Delta_t\leftarrow J(E_t;x)-J(E_{t-1};x)$
    \IF{$\Delta_t<0$, or this is the second consecutive zero-gain step}
      \STATE Prune the branch
    \ELSE
      \STATE Retain the child and record whether a zero-gain step has already
      been tolerated
    \ENDIF
  \ENDFOR
  \STATE Keep at most the top $K$ children by committee score
\ENDFOR
\STATE Generate final answers from surviving states
\RETURN Trajectories whose final answers match $y$
\end{algorithmic}
\vspace{0.4em}
\hrule
\caption{Capability-aligned construction of SFT trajectories. Candidate steps
are executed before scoring, and only non-regressive branches with valid tool
returns can survive.}
\label{alg:supp_sft}
\end{figure*}

\subsection{The Distilled SFT Baseline}

The Distilled SFT baseline used in the main text distills trajectories from
Qwen3.6-plus on the same question pool as ToolVision SFT, with the same
visual toolbelt exposed through the same code interface. Tool use is at the
teacher's discretion: it may invoke tools or answer directly. Trajectories
are retained whenever they execute successfully and end in a correct final
answer, without committee filtering, yielding $4{,}257$ trajectories,
comparable to the $4{,}057$ of ToolVision SFT. Because the teacher's own perception rarely needs specialist
localization, the retained trajectories contain almost no detector-guided
cropping---the pattern the student then imitates (Figure~\ref{fig:dist}).

\subsection{MUT Measurement and RL-Set Construction}
\label{sec:supp_mut}

The tool-enabled arm permits both tool calls and direct answers. The grounded
tool-correct count (GTC) counts correct trajectories that choose to invoke at
least one tool, while the no-tool correct count (NTC) provides the
direct-answer counterfactual.

\begin{figure*}[!t]
\small
\hrule
\vspace{0.5em}
\noindent\textbf{Algorithm 2: MUT measurement and RL-set construction}
\begin{algorithmic}[1]
\REQUIRE Frozen SFT policy $\pi_{\mathrm{SFT}}$ and public question pool
$\mathcal{Q}$
\FOR{each question $x\in\mathcal{Q}$}
  \STATE Run $16$ no-tool rollouts and count correct answers $c(x)$
  \IF{$c(x)\leq8$}
    \STATE Run $8$ tool-enabled rollouts; compute GTC and NTC
    \IF{GTC$\geq2$ and NTC$\leq1$}
      \STATE Assign strong MUT with $w(x,\pi_{\mathrm{SFT}})=0.5$
    \ELSIF{GTC$=1$}
      \STATE Assign weak MUT with $w(x,\pi_{\mathrm{SFT}})=0.2$
    \ENDIF
  \ELSIF{$9\leq c(x)\leq15$}
    \STATE Add $x$ to the source-aware ordinary pool with
    $w(x,\pi_{\mathrm{SFT}})=0$
  \ENDIF
\ENDFOR
\STATE Source-balance the ordinary pool and merge the three groups
\RETURN Fixed RL set and confidence weights $w(x,\pi_{\mathrm{SFT}})$
\end{algorithmic}
\vspace{0.4em}
\hrule
\caption{Paired measurement of tool necessity and construction of the fixed
RL mixture. The tool-enabled arm allows either tool use or a direct answer.}
\label{alg:supp_mut}
\end{figure*}

For an RL trajectory $\xi$, $R_{\mathrm{mut}}(\xi)=1$ only when the final
answer is correct, at least one tool is used, and no tool execution fails. We
use the same symbolic reward definition as in the main text:
\[
\begin{aligned}
R ={}& R_{\mathrm{acc}}+\lambda R_{\mathrm{proto}}
+w(x,\pi_{\mathrm{SFT}})R_{\mathrm{mut}} \\
&-\mu\max(0,n-\tau),
\end{aligned}
\]
where $R_{\mathrm{acc}}$ indicates final-answer correctness,
$R_{\mathrm{proto}}$ checks response and tool-call format, and $n$ is the
number of tool calls. The final setting uses $\lambda=0.2$, $\mu=0.05$,
$\tau=6$, and confidence weights $w=0.5/0.2/0$ for strong MUT, weak MUT, and
ordinary questions, respectively.

\section{Data Sources and Splits}

\subsection{SFT and RL Sources}

The SFT set contains $4{,}057$ examples from the public sources summarized in
Table~\ref{tab:supp_sft_sources}. Tool-use trajectories and step-utility
signals are synthesized without additional human tool-use annotations. All
source images, questions, and reference answers retain their original dataset
licenses.

The RL pool combines publicly released sources spanning general and diagram
QA, text and chart understanding, scientific figures, spatial reasoning,
grounding, visual search, and counting. We first apply the paired measurement
described below and then perform source-aware sampling. Table~\ref{tab:supp_rl_sources}
reports the exact composition of the resulting $24{,}921$-question set. The
three columns correspond to the frozen confidence tiers used by training:
$w=0.5$ for strong MUT, $w=0.2$ for weak MUT, and $w=0$ for ordinary
questions.

\begin{table*}[!t]
\centering
\small
\setlength{\tabcolsep}{6pt}
\begin{tabular}{@{}p{0.27\textwidth}r p{0.46\textwidth}@{}}
\toprule
Source & Examples & Role in SFT \\
\midrule
GQA~\citep{hudson2019gqa} & 1,155 & General visual QA \\
CA-VQA~\citep{daxberger2025mmspatial} & 793 & Spatial and attribute QA \\
TextVQA~\citep{singh2019textvqa} & 598 & Text reading \\
FSC-147~\citep{ranjan2021fsc147} & 511 & Counting \\
CodeVision~\citep{guo2026codevision} & 1,000 & Rotate/flip seeding \\
\midrule
Total & 4,057 & \\
\bottomrule
\end{tabular}
\caption{Composition of the SFT set. The first four sources use their public
training data; the CodeVision examples are from its released training set and
are used to seed rotate and flip operations.}
\label{tab:supp_sft_sources}
\end{table*}

\begin{table*}[!t]
\centering
\small
\setlength{\tabcolsep}{4.5pt}
\begin{tabular}{llrrrr}
\toprule
Source & Released split or subset & Strong & Weak & Ordinary & Total \\
\midrule
GQA~\citep{hudson2019gqa} & train & 961 & 600 & 1,635 & 3,196 \\
TextVQA~\citep{singh2019textvqa} & train & 659 & 183 & 1,047 & 1,889 \\
FSC-147~\citep{ranjan2021fsc147} & validation & 578 & 28 & 240 & 846 \\
ChartQA~\citep{masry2022chartqa} & train & 154 & 323 & 872 & 1,349 \\
ArxivQA~\citep{li2024arxivqa} & public pool, excluding ArxivQA-2k & 357 & 1,807 & 661 & 2,825 \\
PixMo-Count~\citep{deitke2024molmo} & train & 1,181 & 446 & 218 & 1,845 \\
ViGoRL visual search~\citep{sarch2025vigorl} & released visual-search set & 1,277 & 1,256 & 1,563 & 4,096 \\
ViGoRL SAT-2~\citep{sarch2025vigorl} & released SAT-2 set & 1,768 & 661 & 859 & 3,288 \\
Ref-L4~\citep{chen2025refl4} & public release & 575 & 233 & 1,201 & 2,009 \\
AI2D~\citep{kembhavi2016ai2d} & train via The Cauldron & 71 & 320 & 646 & 1,037 \\
CountQA~\citep{tamarapalli2025countqa} & public test split & 299 & 140 & 192 & 631 \\
MMStar~\citep{chen2024mmstar} & validation & 101 & 264 & 239 & 604 \\
InfographicVQA~\citep{mathew2022infographicvqa} & validation & 215 & 125 & 197 & 537 \\
OCRBench v2~\citep{fu2025ocrbenchv2} & public test, text-recognition subset & 174 & 54 & 221 & 449 \\
DocVQA~\citep{mathew2021docvqa} & validation & 90 & 41 & 189 & 320 \\
\midrule
Total & & 8,460 & 6,481 & 9,980 & 24,921 \\
\bottomrule
\end{tabular}
\caption{Exact source distribution of the final RL set after paired
measurement and source-aware sampling. Strong, weak, and ordinary denote the
three fixed necessity-confidence tiers used during RL. CountQA and OCRBench v2
are public benchmark splits used only as RL sources; neither is used as a
reported evaluation set in this paper.}
\label{tab:supp_rl_sources}
\end{table*}

\subsection{ArxivQA-2k Holdout}

Before preprocessing the training pool, we randomly sample and fix $2{,}000$
ArxivQA questions for evaluation. Their question identifiers are excluded from
the SFT and RL mixtures, so the reported ArxivQA-2k result is question-disjoint
from training. This statement concerns question-level isolation and does not
assume that different questions never reuse the same source figure.

\section{Evaluation Protocol}

Benchmark instances and metric implementations follow LMMs-Eval where
available~\citep{zhang2024lmmseval}. V\textsuperscript{*}, HRBench 4K/8K,
MME-RealWorld-Lite/CN, and ArxivQA-2k use multiple-choice option accuracy.
ChartQA uses relaxed accuracy with a $5\%$ tolerance for numerical answers;
OCRBench uses inclusion-based text matching; and FSC-147-test uses mean
absolute error (MAE; lower is better). Following the released CodeVision
evaluation procedure~\citep{guo2026codevision}, a fixed, zero-temperature,
reference-conditioned LLM judge is used only when a task rule cannot reliably
resolve semantic equivalence.
At evaluation, we sample four trajectories per question at temperature $0.7$
and report avg@4 as the primary accuracy-oriented metric.
Table~\ref{tab:supp_std} reports the standard deviation of ToolVision-8B's
scores across the four evaluation runs, where each run scores one of the four
sampled trajectories per question.

\begin{table*}[!t]
\centering
\small
\begin{tabular}{lccccccccc}
\toprule
 & V\textsuperscript{*} & HR-4K & HR-8K & OCRBench & ChartQA & MME-L & MME-CN
 & ArxivQA-2k & FSC-147 \\
\midrule
Std of per-run scores & 0.79 & 0.99 & 0.50 & 0.31 & 0.25 & 0.58 & 0.17 & 0.33
& 0.67 \\
\bottomrule
\end{tabular}
\caption{Standard deviation of ToolVision-8B benchmark scores across the four
evaluation runs underlying avg@4. FSC-147 is in MAE points; all other columns
are accuracy points.}
\label{tab:supp_std}
\end{table*}

\section{Safeguard Activation}

The protocol and overuse terms prevent malformed responses and extreme tool
overuse rather than driving the task reward. Across the complete 194-step
training log, which includes the continuation beyond the step-180 checkpoint
reported in the paper, $R_{\mathrm{proto}}=1$ for $99.72\%$ of the $198{,}656$
logged trajectories. The overuse penalty activates on only $297$ trajectories
($0.15\%$), whose tool-call counts exceed $\tau=6$. These low activation rates
show that both terms act as safeguards for exceptional failures.

\section{Task--Tool Specialization}

Table~\ref{tab:supp_tooldist} reports, for representative benchmarks, the
share of ToolVision-8B evaluation questions on which the semantically matched
specialist operation is invoked at least once, computed on complete
evaluation generations. Task--tool alignment is sharp after RL: grounding
dominates high-resolution search, OCR dominates text reading, and counting
saturates FSC-147. Manual coordinate cropping is essentially absent
($0.0\%$ macro average over the nine benchmarks; at most $0.1\%$ on any
single one), and unmodified-image re-display stays below $2\%$ of questions
on every benchmark.

As a tool-only reference for FSC-147-test, invoking the counting tool alone
on every image yields an MAE of $14.76$, above ToolVision-8B's $11.56$: the
policy adds value beyond routing questions to the specialist.

\begin{table}[!t]
\centering
\small
\begin{tabular}{llc}
\toprule
Benchmark & Operation & Share (\%) \\
\midrule
V\textsuperscript{*} & Grounding & 71.7 \\
OCRBench & OCR & 84.2 \\
FSC-147-test & Counting & 100.0 \\
\bottomrule
\end{tabular}
\caption{Share of ToolVision-8B evaluation questions that invoke the
task-matched specialist operation at least once.}
\label{tab:supp_tooldist}
\end{table}

\section{Prompt and Output Interfaces}

At inference and during RL, the policy sees exactly one system prompt and
one tool contract; the Planner--Executor--Judge structure of
Algorithm~1 exists only inside the SFT synthesis pipeline,
where each role receives a separate minimal interface. This section
reproduces those interfaces and states their information boundaries.

\subsection{Policy System Prompt}

The same system prompt is used for SFT training targets, RL rollouts, and
evaluation; it is reproduced in Figure~\ref{fig:supp_sysprompt}. It fixes the
response protocol (\texttt{think}, then either a tool call or an answer) and
gives evidence-oriented guidance for choosing operations. It contains no
task-specific hints and no information about rewards, judges, or reference
answers.

\begin{figure*}[!t]
\hrule
\vspace{0.45em}
\footnotesize
\begin{verbatim}
You are an advanced MLLM that can solve complex problems with image processing and
analysis tools. You must think step-by-step inside the <think></think> tags first to
determine the next action. If tools are needed to inspect evidence, call them inside
the <tool_call></tool_call> tags. Otherwise, provide your final answer within the
<answer></answer> tag.

## Analysis Chain

**Think and Analyze** -> **Use Evidence Tools When Helpful** -> **Validate**
-> **Iterate When Needed** -> **Answer**

1. **Think and Analyze**: Examine the image and question, identify key objects, text,
   regions, counts, spatial relations, charts, or other visual evidence needed for the
   answer.
2. **Use Evidence Tools When Helpful**: Select the tool operation that best matches the
   missing evidence. Use OCR for text-heavy or hard-to-read text, counting assistance
   for object-counting questions, crop or zoom for small details or local regions,
   point or line drawing for precise locations, boundaries, alignments, or comparisons,
   grounding or object-focused crop for locating described objects, depth tools for
   relative-depth questions, and other image operations only when they directly help
   inspect the evidence.
3. **Validate**: Analyze the tool returns and check whether they answer the question.
   Cross-check with the original image or another focused tool call when needed.
4. **Iterate When Needed**: If the tool invocation fails, the observation is
   incomplete, or the current hypothesis is uncertain, analyze the reason and choose a
   better targeted action.
5. **Answer**: Once the visual evidence supports an answer, provide your final answer
   inside the <answer></answer> tag.

**Key Principles**: Be systematic, use tools only to reveal task-relevant evidence,
prefer the simplest targeted operation, and avoid unnecessary image modifications.
\end{verbatim}
\vspace{0.3em}
\hrule
\caption{The policy system prompt, shared by SFT, RL, and evaluation. The
text is verbatim; line breaks are adjusted for typesetting.}
\label{fig:supp_sysprompt}
\end{figure*}

\subsection{Tool Contract}

All operations go through a single function, \texttt{code\_image\_tool},
whose arguments are executable Python \texttt{code}, a natural-language
\texttt{description}, and an \texttt{image\_index} selecting the input from
the visible image timeline (root images first, then each successful tool
return, append-only). The code runs against a pre-bound \texttt{image}
variable; reading from local paths or URLs is disallowed, and the resulting
image must be assigned to \texttt{result}. Helper functions expose the
external visual services; Table~\ref{tab:supp_helpers} groups them by
capability. Syntax errors, runtime errors, tool-service failures, empty
outputs, and invalid outputs are all rejected: the call returns an error
message instead of an image, and the turn simply continues.

\begin{table}[!t]
\centering
\footnotesize
\begin{tabular}{@{}ll@{}}
\toprule
Capability & Helper functions \\
\midrule
OCR & \texttt{\_call\_ocr\_assist} \\
Counting & \texttt{\_call\_count\_assist} \\
Localization\,/\,crop & \texttt{\_call\_ground\_box}, \\
& \texttt{\_call\_dino\_crop}, \\
& \texttt{\_call\_manual\_box}, \\
& \texttt{\_call\_manual\_crop} \\
Segmentation focus & \texttt{\_call\_sam\_mask}, \texttt{\_call\_blur\_bg} \\
Depth & \texttt{\_call\_manual\_depth}, \\
& \texttt{\_call\_ground\_depth} \\
Direct image ops & PIL/OpenCV/NumPy (rotate, flip, \\
& brightness, contrast, crop, \dots) \\
\bottomrule
\end{tabular}
\caption{Helper families available inside \texttt{code\_image\_tool}. The
prompt documents each helper with one usage line and a one-sentence
applicability rule.}
\label{tab:supp_helpers}
\end{table}

\subsection{Turn-Level Feedback Template}

Every tool return is wrapped in the same continuation template, shown in
Figure~\ref{fig:supp_feedback}. Successful and failed calls use the same
template (failures replace the first sentence with the error message), so
the decision to keep calling tools or to answer always remains with the
policy. No variant of this template comments on whether the current
reasoning is on the right track.

\begin{figure}[!t]
\hrule
\vspace{0.45em}
\footnotesize
\begin{verbatim}
Here is the processed image. Now, analyze
the returned results. Please keep thinking
step-by-step inside the <think></think>
tags to determine the next action. If
additional tools are required, call them
inside the <tool_call></tool_call> tags.
Otherwise, provide your final answer
within the <answer></answer> tags.
\end{verbatim}
\vspace{0.1em}
\footnotesize On failure, the leading sentence is replaced by the error
text, e.g.\ \texttt{Error: The result has an invalid image size (0x136).},
followed by the same continuation.
\vspace{0.35em}
\hrule
\caption{The turn-level feedback template appended to every tool response.}
\label{fig:supp_feedback}
\end{figure}

\subsection{Synthesis-Time Interfaces}

\paragraph{Planner.}
The Planner receives the conversation so far (question, visible images,
prior actions, and tool outputs), the visible-image timeline, the remaining
step budget, and the available capability list, and must return exactly one
JSON object: either \texttt{mode="answer"} or \texttt{mode="suggestions"}
with at most two executable strategy branches. Which modes are legal in a
given round is controlled by a three-level round policy:
\texttt{MUST\_SUGGEST} (opening rounds), \texttt{MAY\_ANSWER\_OR\_SUGGEST}
(intermediate rounds), and \texttt{MUST\_ANSWER} (final round, once the
execution budget is exhausted). Figure~\ref{fig:supp_roundpolicy} reproduces
the \texttt{MUST\_ANSWER} block, which is the only mechanism that ever
forces an answer: it closes the trajectory but supplies no answer content
beyond the committee's own consensus candidate, derived from visible
evidence and re-verified against the image.

\begin{figure}[!t]
\hrule
\vspace{0.45em}
\footnotesize
\begin{verbatim}
Round policy:
- This round is `MUST_ANSWER`.
- Return `mode="answer"`.
- Do not return any `suggestions` field.
- Finalize the best answer now from the
  visible evidence and executed trajectory
  above.
- The executed trajectory appears
  sufficient; answer from the visible
  evidence without citing any hidden
  policy or evaluation signal.
- Independent judge consensus from the
  executed trajectory proposes this final
  answer: `<consensus candidate>`. Treat
  it as a high-confidence candidate from
  the visible/tool evidence; verify it
  against the visible image before
  finalizing.

Budget constraints:
- `remaining_exec_steps = <k>` is the
  total number of executor steps still
  available on this trajectory before the
  final answer.
- Every suggested branch must fit within
  this remaining budget.
- When the executable budget is
  exhausted, the caller will switch to a
  final-answer round.
\end{verbatim}
\vspace{0.3em}
\hrule
\caption{The final-round (\texttt{MUST\_ANSWER}) policy block and the budget
block of the Planner user prompt (verbatim; the last two policy lines are
appended only when the trajectory already executed successfully and when a
committee consensus exists, respectively).}
\label{fig:supp_roundpolicy}
\end{figure}

\paragraph{Executor.}
The Executor turns one suggested step into one concrete tool call under a
strict single-step contract: produce a step rationale and exactly one
\texttt{code\_image\_tool} call whose \texttt{think}, code, and description
agree; never output a final answer; start from the pre-bound input image
rather than files or URLs; and preserve the most useful resulting image in
\texttt{result}. Rejected executions (syntax, runtime, tool-service,
empty-output, or invalid-output failures) terminate the branch rather than
being papered over.

\paragraph{Committee Judge.}
Each of the eight committee members receives the instruction in
Figure~\ref{fig:supp_judge} together with the executed trajectory and its
images. Judges answer the original question independently; a task-specific
matcher outside the generation loop then compares each returned answer with
the reference to produce the normalized step score that drives branch
pruning.

\begin{figure}[!t]
\hrule
\vspace{0.45em}
\footnotesize
\begin{verbatim}
You are a multimodal judge model.

Your job is to solve the original
question from the executed trajectory and
visible evidence.

Rules:
- Use the conversation history and
  attached images as evidence.
- Do not explain your reasoning.
- Return only the final answer text.
\end{verbatim}
\vspace{0.3em}
\hrule
\caption{The complete committee-judge system prompt. The reference answer is
not part of the judge input.}
\label{fig:supp_judge}
\end{figure}

\subsection{What the Interfaces Never Contain}

No generation-side prompt---policy, Planner, Executor, or Judge---contains
the reference answer at any stage. Judges solve the question independently
from the trajectory evidence, and the reference enters only the external
matcher that scores their returned answers. The consensus candidate
optionally shown to the Planner in the final round
(Figure~\ref{fig:supp_roundpolicy}) is an aggregate of those independent
judge answers, not an annotation, and the Planner is explicitly instructed
to verify it against the visible image. The same boundary holds for the MUT
measurement of Algorithm~2, whose two arms reuse the policy
interface unchanged with tool access disabled or enabled. This implements
the claim in the main text that both supervision signals are constructed
automatically from public task data without additional human annotation of
tool use or necessity, and, beyond that, without exposing reference answers
to any trajectory generator.

\begin{figure*}[!t]
\centering
\begin{minipage}[c]{0.385\textwidth}
\centering
\includegraphics[width=\textwidth]{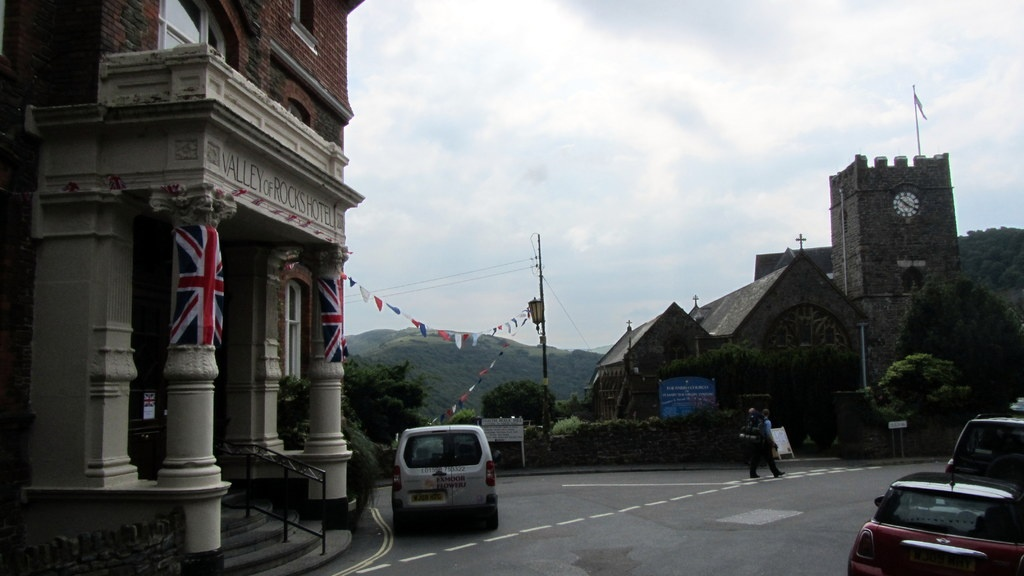}\par
{\footnotesize (a) input, 1024$\times$576}
\end{minipage}\hspace{0.7em}%
\begin{minipage}[c]{0.085\textwidth}
\centering
\fbox{\includegraphics[width=0.8\textwidth]{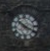}}\par
{\footnotesize (b) crop, 50$\times$51}
\end{minipage}\hspace{0.7em}%
\begin{minipage}[c]{0.185\textwidth}
\centering
\includegraphics[width=0.92\textwidth]{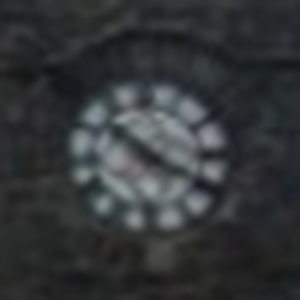}\par
{\footnotesize (c) resized, 300$\times$300}
\end{minipage}\hspace{0.7em}%
\begin{minipage}[c]{0.185\textwidth}
\centering
\includegraphics[width=0.92\textwidth]{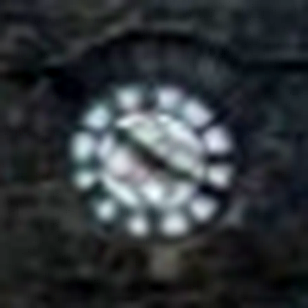}\par
{\footnotesize (d) sharpened + contrast}
\end{minipage}
\vspace{0.45em}
\hrule
\vspace{0.5em}
\begin{minipage}{0.985\textwidth}
\small
\textbf{Question.}\; what time is the clock at?

\medskip
\textbf{Turn 1 --- think.}\; \emph{The user wants to know the time displayed
on the clock tower visible on the right side of the image. The clock face is
present but too small to read the hands accurately in the current full image.
To solve this, I need to isolate and enlarge the clock face. I will use the
\textup{\texttt{dino\_crop}} tool with the text prompt `clock face' to
automatically locate and crop the clock tower's clock. This will give me a
close-up view to read the time.}

\textbf{Turn 1 --- tool call.}\;
\texttt{crop = \_call\_dino\_crop(text\_prompt='clock face', max\_crops=1,
padding=10)}

\textbf{Tool response.}\; \texttt{Image size = 50x51 pixels.
GroundedSAM2(dino\_crop) returned 1 crop images.}

\medskip
\textbf{Turn 2 --- think.}\; \emph{The current evidence is a tiny 50x51 pixel
crop of the clock face, which is insufficient to read the time. The hands and
numerals are indistinct at this resolution. To resolve the uncertainty about
the time shown, I need to magnify this crop. I will resize the current image
to 300x300 pixels, which will make the clock face large enough for visual
inspection and allow me to determine the position of the hands.}

\textbf{Turn 2 --- tool call} (on image 1).\;
\texttt{result = image.resize((300, 300), Image.LANCZOS)}

\textbf{Tool response.}\; \texttt{Image size = 300x300 pixels. Code executed
successfully}

\medskip
\textbf{Turn 3 --- think.}\; \emph{The current image is a 300x300 crop of the
clock face, but the details are blurry and the contrast is low, making the
hands hard to distinguish. To read the time, I need to improve the clarity of
the clock face. Applying sharpening will help define the edges of the hands
and markers. Increasing the contrast will separate the dark hands from the
lighter background. I will apply both enhancements to the current image to
make the time legible.}

\textbf{Turn 3 --- tool call} (on image 2).\;
\texttt{sharp\_image = ImageEnhance.Sharpness(image).enhance(1.8);
result = ImageEnhance.Contrast(sharp\_image).enhance(1.8)}

\textbf{Tool response.}\; \texttt{Image size = 308x308 pixels. Code executed
successfully}

\medskip
\textbf{Turn 4 --- think.}\; \emph{The executed trajectory successfully
located and enhanced the clock face. The original image (index 0) showed a
distant clock on a church tower. A crop was taken (index 1), resized (index
2), and then sharpened/contrasted (index 3). The final enhanced image (index
3) clearly shows the clock hands. The hour hand (shorter) points towards the
10 position (upper-left quadrant). The minute hand (longer) points towards
the 4 position (lower-right quadrant, representing 20 minutes). Therefore,
the time is approximately 10:20.}

\textbf{Answer.}\; 10:20
\end{minipage}
\vspace{0.5em}
\hrule
\caption{Case 1 (SFT training example). The recorded intermediate images
(a)--(d) are the model's visible image timeline. The clock face is illegible
at full-image scale and still illegible in the raw $50\times51$ detector
crop; the trajectory keeps transforming the evidence---resize, then sharpen
and contrast---until the hands can be read. Model text is verbatim; code and
tool responses are excerpted.}
\label{fig:supp_case1}
\end{figure*}

\begin{figure*}[!t]
\centering
\begin{minipage}[c]{0.24\textwidth}
\centering
\includegraphics[width=\textwidth]{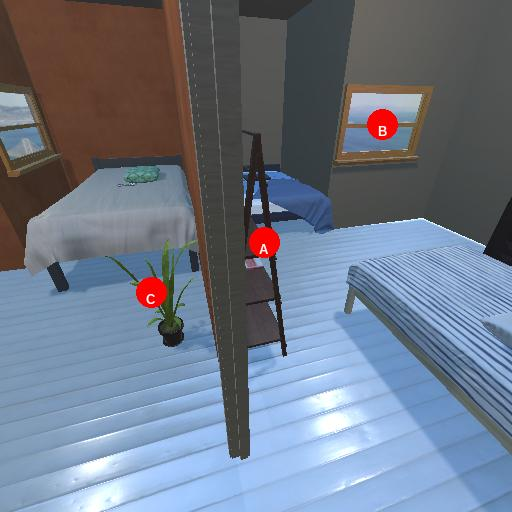}\par
{\footnotesize (a) input, 504$\times$504}
\end{minipage}\hspace{1.0em}%
\begin{minipage}[c]{0.24\textwidth}
\centering
\includegraphics[width=\textwidth]{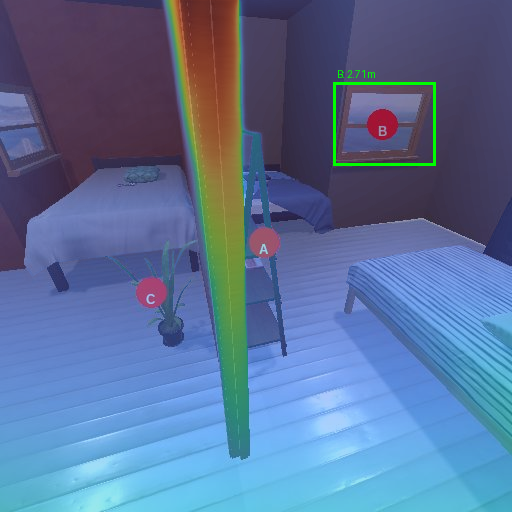}\par
{\footnotesize (b) B grounded: 2.71\,m}
\end{minipage}\hspace{1.0em}%
\begin{minipage}[c]{0.24\textwidth}
\centering
\includegraphics[width=\textwidth]{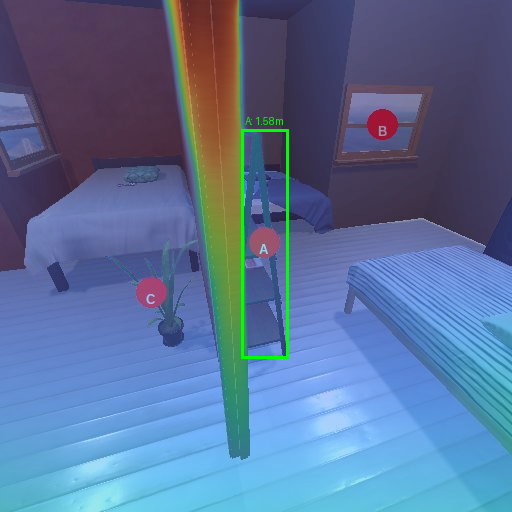}\par
{\footnotesize (c) A grounded (replayed)}
\end{minipage}
\vspace{0.45em}
\hrule
\vspace{0.5em}
\begin{minipage}{0.985\textwidth}
\small
\textbf{Question} (SAT-2).\; Is black color shelving unit (marked A) behind
Window (marked B)?

\medskip
\textbf{Turn 1 --- think.}\; \emph{The image shows two labeled objects: a
red-circle-marked window `B' on the right wall and a red-circle-marked object
`A' near a ladder in the center. The question asks whether the shelving unit
(A) is behind the window (B). To determine depth, I need to estimate the
distance of object A from the camera. I will use the
\textup{\texttt{ground\_depth}} helper to ground the object described as
`black shelving unit' and measure its depth. This will provide the depth
value for A, which can then be compared with B's depth in a subsequent step
to answer the question.}

\textbf{Turn 1 --- tool call.}\;
\texttt{depth = \_call\_ground\_depth("black shelving unit",
detection\_index=0, stat="median", label="A")}

\textbf{Tool response.}\; \texttt{No objects found for 'black shelving
unit.'.}

\medskip
\textbf{Turn 2 --- think.}\; \emph{The previous attempt to ground the `black
shelving unit' failed, likely due to the specific visual appearance of the
object (wooden ladder-like structure with a black base) not matching the
generic descriptor. To resolve the spatial relationship between A and B, I
need to establish a baseline depth for the window B. This will allow me to
compare the depth of A against B in a subsequent step. I will now use the
\textup{\texttt{ground\_depth}} helper with the more concrete description
`window marked B' to measure its depth.}

\textbf{Turn 2 --- tool call.}\;
\texttt{depth = \_call\_ground\_depth("window marked B",
detection\_index=0, stat="median", label="B")}

\textbf{Tool response.}\; \texttt{Detected 3 objects for 'window marked b.';
using detection 0. B median depth is 2.71m.}

\medskip
\textbf{Turn 3 --- think.}\; \emph{The depth of window B has been established
at 2.71m. To determine if the shelving unit A is behind it, I need the depth
of A. The previous attempt to ground `black shelving unit' failed. I will now
try a more specific grounding prompt `ladder with black base' to locate A and
measure its depth. This should help establish whether A is closer to the
camera than B or further away, which directly answers the spatial
relationship question.}

\textbf{Turn 3 --- tool call.}\;
\texttt{depth = \_call\_ground\_depth("ladder with black base",
detection\_index=0, stat="median", label="A")}

\textbf{Tool response.}\; \texttt{Detected 1 objects for 'ladder with black
base.'; using detection 0. A median depth is 1.57m.}

\medskip
\textbf{Turn 4 --- think.}\; \emph{The depth measurements are sufficient to
answer the question. The window B has a depth of 2.71m, while the shelving
unit A (ladder with black base) has a depth of 1.57m. Since 1.57m $<$ 2.71m,
object A is closer to the camera than object B. Therefore, A is not behind
B.}

\textbf{Answer.}\; no \hfill (reference: no; correct)
\end{minipage}
\vspace{0.5em}
\hrule
\caption{Case 2 (RL rollout, SAT-2). The replayed depth visualizations show
the grounded boxes and median depth estimates that the model compared. Model
text is verbatim; code and tool responses are excerpted.}
\label{fig:supp_case2}
\end{figure*}

\begin{figure*}[!t]
\centering
\begin{minipage}[c]{0.365\textwidth}
\centering
\includegraphics[width=\textwidth]{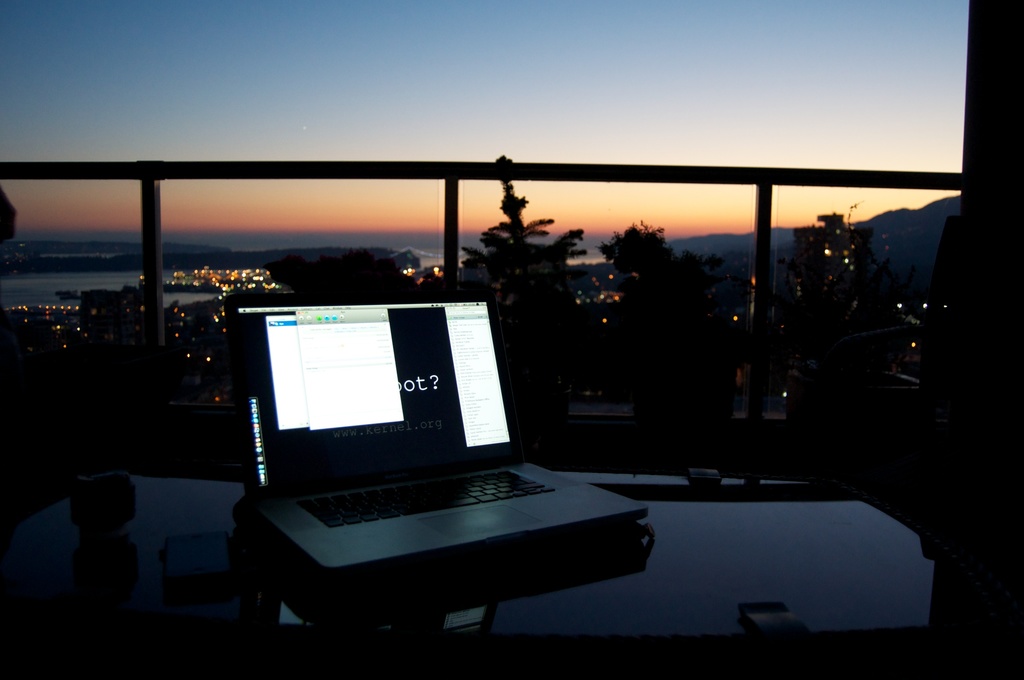}\par
{\footnotesize (a) input}
\end{minipage}\hspace{1.0em}%
\begin{minipage}[c]{0.115\textwidth}
\centering
\fbox{\includegraphics[width=0.9\textwidth]{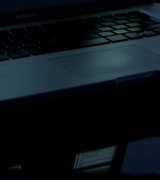}}\par
{\footnotesize (b) manual crop 1}
\end{minipage}\hspace{1.0em}%
\begin{minipage}[c]{0.24\textwidth}
\centering
\includegraphics[width=\textwidth]{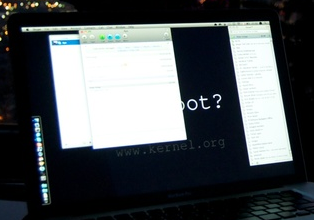}\par
{\footnotesize (c) \texttt{dino\_crop} result}
\end{minipage}
\vspace{0.45em}
\hrule
\vspace{0.5em}
\begin{minipage}{0.985\textwidth}
\small
\textbf{Question} (TextVQA).\; what kind of mark is shown on the black part
of the screen?

\medskip
\textbf{Turn 1 --- think.}\; \emph{The user is asking about the mark on the
black part of the laptop screen. From the image, there appears to be a dark
rectangular area on the right side of the screen where some white text is
visible, likely reading something like `bot?'. To clarify what the mark says,
I will crop the specific region on the right side of the screen where the
text is located. Based on visual estimation, this text is situated roughly in
the center-right area of the screen, so I will crop a region around that area
to isolate the text for better readability.}

\textbf{Turn 1 --- tool call (excerpt).}\;
\texttt{xmin, ymin, xmax, ymax = 370, 460, 530, 640}\,
(clamped to image bounds)\,
\texttt{result = image.crop((xmin, ymin, xmax, ymax))}

\textbf{Tool response.}\; \texttt{Image size = 160x180 pixels. Code executed
successfully}

\medskip
\textbf{Turns 2--3.}\; The crop executed, but it landed on the keyboard
(panel~b): \emph{``The previous crop missed the screen area entirely,
focusing on the keyboard instead.''} Two further manual windows
(390--560, then 220--620 horizontally) also execute successfully and still
return keyboard and table regions ($170\times180$ and $400\times232$).

\medskip
\textbf{Turn 4 --- think.}\; \emph{The previous manual crops failed to
isolate the text because they targeted the keyboard area instead of the
screen. To solve this, I will use \textup{\texttt{dino\_crop}} with the
prompt `text on the screen' to automatically locate and crop the relevant
region. This should extract the small white text visible on the right side
of the laptop screen, allowing me to read the mark clearly.}

\textbf{Turn 4 --- tool call.}\;
\texttt{crop = \_call\_dino\_crop(text\_prompt='text on the screen',
max\_crops=1, padding=10)}

\textbf{Tool response.}\; \texttt{Image size = 318x218 pixels.
GroundedSAM2(dino\_crop) returned 1 crop images.}

\medskip
\textbf{Turn 5 --- think.}\; \emph{The current evidence is sufficient. The
previous step successfully cropped the laptop screen region (image index 3),
clearly showing the text on the black part of the screen. The text visible
is `bot?' and below it `www.kernel.org'. The question asks for the `mark',
which refers to the text. Since the visible text ends with a question mark,
the mark is a question mark.}

\textbf{Answer.}\; question mark \hfill (reference: \texttt{?}; scored
correct by the benchmark's answer matcher)
\end{minipage}
\vspace{0.5em}
\hrule
\caption{Case 3 (RL rollout, TextVQA). All three manual crops execute without
error yet land on the keyboard rather than the screen; the text-conditioned
detector crop isolates the on-screen text in one step. Model text is
verbatim; code and tool responses are excerpted.}
\label{fig:supp_case3}
\end{figure*}

\section{Qualitative Case Studies}

All cases below are recorded trajectories, not curated demonstrations: the
model text is reproduced verbatim (code and tool observations are excerpted
where marked), and placeholders in angle brackets substitute run-specific
values in quoted prompts. Case~1 is an example from the released SFT set,
shown with its recorded intermediate images; Cases~2 and~3 are RL rollouts
collected during training, whose intermediate images are replays of the
recorded code on the deployed visual services and can therefore differ from
the original returns at the pixel level.

\subsection{Case 1: Making a Distant Clock Readable (SFT)}

Case~1 (Figure~\ref{fig:supp_case1}) shows what a capability-aligned SFT
target looks like. The chain is the shortest route from unreadable to
readable evidence: a text-conditioned detector isolates a clock face that
occupies roughly $50\times51$ of the $1024\times576$ input, resizing
magnifies it, and sharpening plus contrast separate the hands from the dial.
No step answers the question by itself, and no step is decorative; each
strictly increases the evidence available for the final readout (hour hand
at 10, minute hand at 4), stated in the reasoning before the answer is
produced. Chains of this shape are what survive the execution checks and the
committee's stepwise evidence-gain pruning of Algorithm~1.

\subsection{Case 2: Depth Verification of a Spatial Claim (RL)}

Case~2 (Figure~\ref{fig:supp_case2}) turns a qualitative spatial question
into an explicit numerical comparison. Monocular layout is misleading
here---A and B sit on opposite sides of a partition---so the model reaches
for the one capability that measures rather than guesses. The first
grounding query fails; instead of guessing, the model fixes a reference by
measuring B, then retries A under a more concrete description (``ladder with
black base'') that the detector can resolve. Two median depths later the
answer is a one-line comparison, $1.57\,\mathrm{m}<2.71\,\mathrm{m}$, and
each number is traceable to a grounded box in the returned visualizations.
The replayed visualization in panel (c) reproduces the recorded measurement
to within a centimeter (1.58\,m versus 1.57\,m).

\subsection{Case 3: Misplaced Manual Crops and Detector Recovery (RL)}

Case~3 (Figure~\ref{fig:supp_case3}) is the failure mode that no error
message ever flags. Every manual crop in this rollout is legal and executes
cleanly---the failure is purely semantic, three coordinate guesses that all
land on the keyboard instead of the screen, a direct expression of the
pixel-level localization gap discussed in the main text. The model diagnoses
the mismatch from the returned images rather than from any error signal,
then hands localization to the detector, whose text-conditioned query
(``text on the screen'') resolves in one step what three coordinate guesses
could not. At the corpus level this preference is the shift visible in
Figure~\ref{fig:dist}, where ToolVision's tool distribution
concentrates on detector-guided cropping; cropping stays in the repertoire,
but the operation trusted for localization is the one whose success does not
depend on the model's own pixel-coordinate estimates.

\section{Reproducibility Details}

\subsection{Code, Data, and Licenses}
The code and data package prepared for this work contains the complete text records for the 4,057
SFT examples and 24,921 RL questions, their source manifests, representative
20-example image subsets for both stages, and the core trajectory-synthesis,
MUT-reward, training, evaluation, and tool-service launchers.  Original
benchmark images, model checkpoints, cluster logs, and private service files
are not redistributed.  The manifests identify the public-source examples
needed to materialize the remaining images under their original dataset
licenses; the bundled subsets support inspection and smoke tests, while a full
training run requires those public images to be materialized locally. All
derived data will be released publicly upon publication under licenses that
permit free research use.

\subsection{Randomness and Number of Runs}
Because full training is computationally expensive, each reported SFT and RL
configuration is trained once, and each result is obtained from one checkpoint
for that configuration. SFT optimization and source-aware sampling for the MUT
mixture use seed 42. The RL data loader uses seed 1, while vLLM rollout
generation and evaluation rollout generation use seed 0; diagnostic evaluation
subsampling uses seed 42. These values will be preserved in the released
configurations. Distributed GPU execution and remote API-based trajectory
synthesis can nevertheless remain nondeterministic.

\subsection{Computing Environment}
Training runs used the same Alibaba Cloud PAI workspace and custom A100
resource pool. Each host contains two 32-core Intel Xeon Platinum 8369B
(Ice Lake) CPUs, 32 $\times$ 64 GB DDR4-3200 memory, two 480 GB SATA SSDs,
four 3.84 TB NVMe SSDs, one dual-port 25 Gbps Ethernet NIC, four dual-port
100 Gbps RoCE NICs, and eight NVIDIA A100-SXM4-80GB GPUs. SFT used one such
eight-GPU host. The final RL job used two workers from the same pool; each
worker was allocated $110$ vCPUs, eight A100-SXM4-80GB GPUs, $1{,}500$ GiB of
host memory, and $1{,}500$ GiB of shared memory. Thus, RL used sixteen
A100-80GB GPUs in total. The jobs ran in an Ubuntu 22.04 container with CUDA
12.9. The activated RL environment used Python 3.11.15, PyTorch 2.8.0,
Transformers 4.57.1, vLLM 0.11.0, Ray 2.54.1, and FlashAttention 2.8.3. SFT
used PyTorch 2.8.0, Transformers 4.57.1, Datasets 4.0.0, and Tokenizers 0.22.2.
Final evaluation jobs used the same PAI workspace, Ubuntu container, CUDA
stack, and activated CodeVision environment as the RL jobs.

The specialist services use PaddleX 3.4.3 with PP-OCRv5-server detection and
recognition and PP-LCNet text-line orientation; GroundingDINO with the Swin-T
OGC checkpoint together with SAM 2.1 Hiera-Tiny; Depth Pro with the released
\texttt{depth\_pro.pt} checkpoint; and CountGD with the FSC-147 ViT-B
configuration, the released \texttt{checkpoint\_fsc147\_best.pth}, and a
BERT-base-uncased text encoder. The GroundingDINO/SAM service uses Python
3.10.20, PyTorch 2.3.1, GroundingDINO 0.1.0, and SAM 2 1.0; the Depth Pro
service uses Python 3.9.23 and PyTorch 2.8.0; the CountGD service uses Python
3.9.19, PyTorch 2.2.1, and Transformers 4.39.1. All specialist services can be
co-located on one additional GPU and use at most 16 GB of GPU memory.

\subsection{Final Training and Evaluation Configurations}
For SFT construction, Qwen3.6-plus serves as Planner and Executor, with Planner
proposal budget $B=2$, beam width $K=2$, and maximum depth $D=6$. The
eight-model committee is fixed to Qwen3.6-plus, Qwen3.5-122B-A10B,
Qwen3.5-35B-A3B, Qwen3-VL at 2B, 4B, and 8B, Gemini 2.5 Flash, and Gemini 2.5
Flash-Lite. Committee scores are equally averaged.

SFT starts from Qwen3-VL-8B-Thinking and trains the language model for two
epochs while freezing the vision tower and multimodal projector. It uses
bfloat16, sequence length 32,768, AdamW with learning rate $10^{-5}$, cosine
decay, 5\% warmup, $\beta_1=0.9$, $\beta_2=0.999$, $\epsilon=10^{-8}$,
per-device batch size 4, gradient accumulation 4, and eight GPUs, giving global
batch size 128. Optimization uses seed 42 and DeepSpeed ZeRO-3.

MUT construction uses 16 no-tool rollouts and, for candidates with at most
eight correct no-tool answers, eight tool-enabled rollouts. The final
$24{,}921$-question mixture contains 8,460 strong, 6,481 weak, and 9,980
ordinary questions with weights 0.5, 0.2, and 0, respectively. GSPO runs for
one epoch with 128 prompts per batch, eight rollouts per prompt, mini-batch size
32, micro-batch size 1 per GPU, and actor learning rate $10^{-6}$. Rollouts use
temperature 0.7, top-$p$ 0.95, maximum prompt and response lengths of 16,384,
maximum tool return length 10,240, maximum per-turn response length 2,048, and
at most 12 assistant turns. vLLM uses tensor parallel size 4, at most 32
concurrent sequences, GPU-memory utilization 0.7, and prefix caching. Training
uses GSPO sequence-level aggregation, clipping coefficients 0.20/0.28 with
$c=10$, KL-loss coefficient 0.001, zero entropy coefficient, FSDP parameter and
optimizer offloading, and gradient checkpointing. Checkpoints are saved every
10 steps, and we report the checkpoint with the highest training reward,
which occurs at step 180.

Two settings were selected by small comparative searches rather than fixed a
priori. The RL prompt batch size was chosen from $\{64, 128\}$: 128 prompts
per step give each batch a more balanced mixture of question sources and
difficulty, which stabilized early training. The vLLM concurrency limit was
decreased from 1{,}024 to 32 concurrent sequences because
Qwen3-VL-8B-Thinking produced malformed tool-call responses under high rollout
concurrency. The remaining optimization settings follow common GSPO practice
and were not tuned.

Evaluation samples four trajectories per question at temperature 0.7 and
reports avg@4. Benchmark-specific rules are applied first; the semantic
equivalence fallback Judge uses zero-temperature decoding.

\subsection{MUT Measurement Cost}
The paired measurement covers $118{,}724$ questions in the no-tool arm (16
rollouts each) and the $51{,}773$ resulting candidates in the tool-enabled arm
(8 rollouts each). On eight-GPU A100-80GB workers, the two arms take $43.2$
and $673.0$ GPU-hours, respectively; the tool-enabled arm dominates because
its trajectories execute live tool services. This one-off $716$ GPU-hour
measurement amounts to roughly $40\%$ of the compute of a single RL run
($1{,}728$ GPU-hours to step 180) and is amortized: the frozen labels are
reused unchanged by every necessity-conditioned RL run.

\end{document}